\documentclass{article}

\usepackage[nonatbib, preprint]{neurips_2026}

\makeatletter
\providecommand{\@notice}{}
\makeatother

\usepackage{amsmath,amsfonts,bm}

\def\eqref#1{equation~\ref{#1}}

\def\1{\bm{1}}

\DeclareMathAlphabet{\mathsfit}{\encodingdefault}{\sfdefault}{m}{sl}
\SetMathAlphabet{\mathsfit}{bold}{\encodingdefault}{\sfdefault}{bx}{n}

\DeclareMathOperator{\Tr}{Tr}

\newcommand{\reals}{\mathbb{R}}

\usepackage[utf8]{inputenc} 
\usepackage[T1]{fontenc}    
\usepackage{url}            
\usepackage{booktabs}       
\usepackage{amsfonts}       
\usepackage{nicefrac}       
\usepackage{microtype}      
\usepackage{xcolor}         
\usepackage{graphicx}
\usepackage{url}
\usepackage{booktabs}
\usepackage{amsmath}

\usepackage{xcolor}
\definecolor{nicecite}{HTML}{2A7FFF} 

\usepackage[
  colorlinks=true,
  linkcolor=black,
  urlcolor=nicecite,
  citecolor=nicecite
]{hyperref} 

\usepackage[
  backend=biber,
  style=numeric-comp,  
  sorting=none,        
  maxcitenames=2,
  maxbibnames=99
]{biblatex}

\usepackage{cleveref}
\crefname{figure}{Fig.}{Figs.}

\usepackage{soul}

\title{How Data Augmentation Shapes Neural Representations}

\author{Tianxiao He 
\\
Department of Computer Science\\
New York University\\
New York, NY 10012 USA \\
\texttt{th3129@nyu.edu} \\
\And
Alex H. Williams \\
Center for Neural Science \\
New York University \\
New York, NY 10012 USA\\
\texttt{aw4614@nyu.edu} \\
\AND
Sarah E. Harvey\\
Flatiron Institute \\
New York, NY 10010, USA \\
\texttt{sharvey@flatironinstitute.org}
}

\begin{document}

\maketitle

\begin{abstract}
Data augmentation is widely recognized for improving generalization in deep networks, yet its impact on the geometry of learned representations remains poorly understood.
In this work, we characterize how different data augmentation strategies reshape internal representations in neural networks. Using tools from shape analysis, we embed network hidden representations into a metric space where distance is invariant to scaling, translation, rotation and reflection. 
We show that increasing augmentation strength leads to well-behaved trajectories in this space, and that different augmentation types steer representations in distinct directions. 
Moreover, we investigate how neural representation shapes are distorted along data augmentation trajectories, and show that insights from neural geometry can predict which representations provide the most improvement when ensembling models.  
Our results reveal shared geometric patterns across architectures and seeds, and suggest that analyzing shape-space trajectories offers a principled tool for understanding and comparing data augmentation methods. Open source code is available at https://anonymous.4open.science/r/netrep-analysis-00EC\end{abstract}

\section{Introduction}
Due to the overparametrized nature of deep neural networks, many hyperparameter settings often achieve similar task performance. 
Tuning hyperparameters such as learning rate, batch size, and momentum is known to be crucial for performance \parencite{smith2018}, but there is little direct study on how these choices change the geometry of learned neural representations.
In particular, when a single hyperparameter is isolated and gradually varied, how do features of hidden layer activity change?

A particularly rich area to study this phenomenon is data augmentation. 
Data augmentation (DA) is a foundational technique for improving the performance of deep networks by diversifying existing datasets with simple transformations \parencite{shorten2019}.  
In the context of computer vision, classic DA techniques include rotation, cropping, color jitter and addition of Gaussian noise to some fraction of input images. 
DA is often interpreted as an implicit regularization technique \parencite{dao2019} or the incorporation of prior knowledge about invariances in the task \parencite{Lenc_2015_CVPR,chen2020}.
Despite their centrality in modern training pipelines, data augmentations are often tuned using simple heuristics or searches that optimize task performance. 
However, many distinct combinations of augmentation hyperparameters, such as strength, probability, or spatial extent, can yield near-identical accuracy on held-out data. 
This degeneracy suggests that performance metrics alone may obscure meaningful differences in how models learn to represent the world under different training protocols.

While the benefits of DA are intuitive, a deeper understanding remains elusive.
For example, is there a sense in which DA methods move representations along well-defined trajectories as the hyperparameter is varied?  
Do distinct augmentation methods (e.g. image rotations and random cropping) sculpt hidden layer computations in fundamentally different ways, or are their effects redundant? 
How do two augmentation methods interact when they are simultaneously applied? 
More broadly, does diversity in representational geometry predict the extent of functional improvements when models are ensembled across different hyperparameter values?  
Can two augmentations synergistically interact or interfere with each other, and can this be observed from their hidden representational geometry?


As a first step towards addressing these questions, we leverage previously developed \textit{metric spaces} on the geometry of hidden layer representations \parencite{Williams2021,Lange2023}. 
This framework identifies the \textit{geometry} of hidden layer representations as points on a smooth manifold of possible geometries, enabling tools from differential geometry to study the effect of hyperparameter changes on learned representations.
In particular, gradual changes in augmentation strength can be conceptualized as moving the representational geometry along a smooth path on this manifold. Different types of augmentation give rise to different paths over the manifold, and  we can compute the lengths and curvatures of these paths, as well as the angles formed at intersections.
Altogether, this enables a systematic comparison of how augmentation strength and types reshape the representation geometry across network layers, initializations, and architectures. Our main contributions are: 

\begin{itemize}
    \item We propose a geometric framework for analyzing the effects of data augmentation hyperparameters on learned neural representations
    \item We demonstrate that increasing data augmentation magnitude steers representations along structured trajectories in this \emph{shape space}, but not in the naive Euclidean space
    \item We study how the magnitude of representation shape change due to data augmentation compares with the magnitude of representation change due to a different random seed
    \item We show that greater divergence between trajectories of different augmentations predicts larger gains from model ensembling
    \item We find that data augmentation displaces shape landmarks non-uniformly, with effects that depend on network depth and data augmentation type.   
\end{itemize}

\section{Related Work}
\label{gen_inst}

\textbf{Data augmentation.}
Data augmentation (DA) is a central technique in modern machine learning pipelines, used to increase data diversity, improve generalization, and enhance robustness \parencite{shorten2019, wang2025,bishop1995, dao2019}.
It is generally thought that augmentations shape neural computation by ``collapsing'' the manifold of representations along nuisance dimensions, so that augmented views of the same input are mapped to similar responses \parencite{Lenc_2015_CVPR}.
The broader principle of view-invariant representations can be traced back decades within the computer vision and visual neuroscience literature \parencite{Olshausen4700,riesenhuber1999hierarchical,dicarlo2007untangling}.

Importantly, the intuition that training with augmentations ``collapses'' augmented views onto a single point only pertains to the representational geometry of the augmented dataset---it does \textit{not} tell us how the manifold of representations over clean images is altered by training with augmentations.
This distinction is particularly important for augmentation techniques that produce out-of-distribution training samples (e.g. adding a large amount of Gaussian noise), with the goal of improving performance of unseen within-distribution samples.
In this paper, we investigate how different augmentations alter representations of non-augmented ``clean images'' in the test set.




\textbf{Representational similarity.}
Comparing neural network representations is essential for understanding training dynamics, transferability, and model equivalence. 
Representational similarity metrics aim to characterize relationships between hidden layer activation patterns across layers and models. 
Metrics like centered kernel alignment (CKA) measure pairwise similarity between activation matrices using kernel-based alignment \parencite{kornblith2019}, and are widely used to study how layer representations change across seeds or architectures \parencite{nguyen2020}. 
More recent work has proposed using formal metric spaces to capture more nuanced geometric structure in representations \parencite{Williams2021} which allows
for more advanced analysis methods on representations such as regression and clustering.
\textcite{Lange2023} leveraged this idea to study the trajectories of representations through layers of deep neural networks, and also introduce concepts from Riemannian geometry to study these paths.
While our work draws on similar methodological ideas, we apply them to study the trajectories of neural representations \emph{across models} as hyperparameters are tuned.  

\section{Methods}\label{sec:methods}

In this section, we define our notion of a metric space for neural representations.
We then discuss geodesics of this metric and how to measure angles between trajectories of representations.
Lastly, we discuss how to interpret the direction of trajectories in this space in terms of \textit{landmark displacements}, which measure how much the hidden layer response changes to each image input after nuisance transformations are removed.  
\begin{figure}[t]
    \centering
    \includegraphics[scale=0.55]{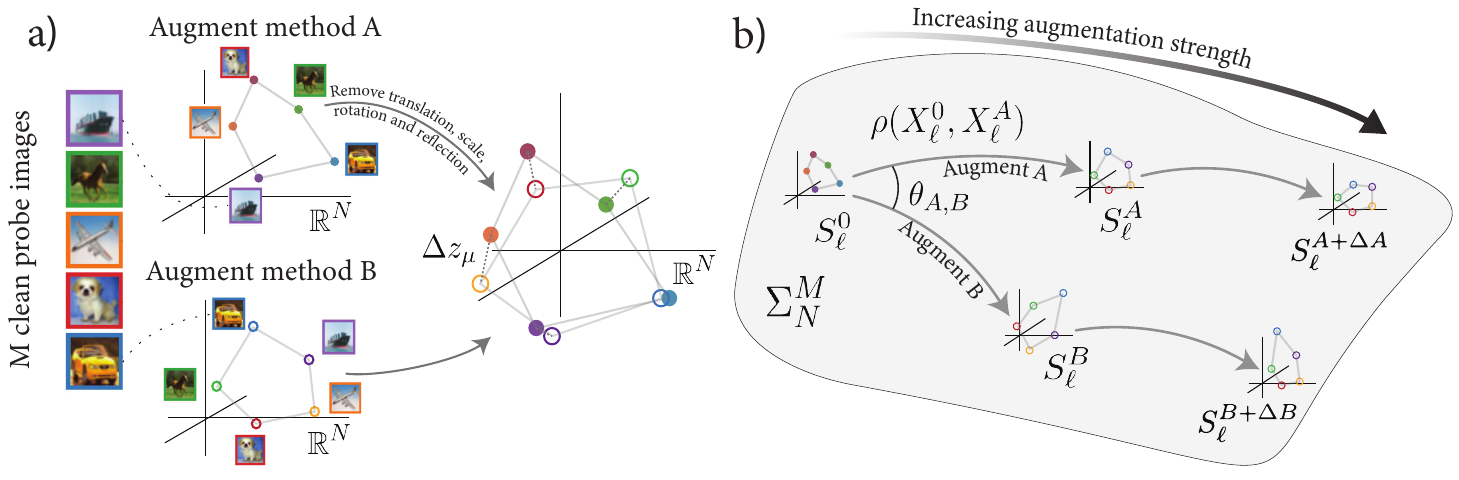} %
  \caption{\textbf{Embedding into shape space enables more meaningful comparison of large ensembles of representations.} \textbf{(a)} After two models are trained end-to-end using different data augmentation methods, $M$ unaltered images from the test dataset are randomly selected as probe images.  The Riemannian shape distance \cref{eq:shape_dist} between these two representations is measured as the angular distance up to translation, scale, rotation, and reflection. \textbf{(b)} The representation shapes from layer $\ell$ with the Riemannian shape distance form a metric space in which we may measure distance, compute geodesics, and measure angles as data augmentation parameters are varied. }\label{fig:cartoon}
\end{figure}

\textbf{Problem Setup.} 
Let $f_L^A: \mathcal{Z} \mapsto \mathcal{Y}$ represent the function of a feedforward neural network with $L \in \mathbb{N}$ layers, where $\mathcal{Z}$ is the input space, $\mathcal{Y}$ is the output space, and the network is trained using DA strategy $A$.
The neural network $f_L^B: \mathcal{Z} \mapsto \mathcal{Y}$ is identical in architecture and initialization to $f_L^A$, but is trained using DA strategy $B$.  

We define the \emph{representation mapping} from the input domain to a vector of activations at a hidden layer $\ell$ with $N$ units as 
$f_\ell^A : \mathcal{Z} \mapsto \mathbb{R}^{N}$ and similarly $f_\ell^B : \mathcal{Z} \mapsto \mathbb{R}^{N}$.
How similar are the activation patterns produced in hidden layers by $f_\ell^A(\cdot)$ and $f_\ell^B(\cdot)$?
More specifically, in what way is the change in training data induced by a DA method change $A \rightarrow B$, or an increase in the magnitude of a hyperparameter $A \rightarrow A + \Delta A$, present in the representations formed by $f_\ell^A(\cdot)$, $f_\ell^B(\cdot)$ and $f_\ell^{A + \Delta A}$?
We measure these changes over a representative collection of unmodified probe inputs $z_1, \hdots, z_M \in \mathcal{Z}$ from the test  set (i.e. clean images).

Following well-established practices for comparing neural representations (see e.g. \cite{kornblith2019,Williams2021}), we proceed by measuring neural responses $f_\ell^A(z_1) \dots f_\ell^A(z_M)$ and stacking them row-wise into a representation matrix $X_\ell^A \in \mathbb{R}^{M \times N}$.
Likewise, we form a matrix $X_\ell^B \in \mathbb{R}^{M \times N}$ from the second network's responses, $f_\ell^B(z_1) \dots f_\ell^B(z_M)$.
Intuitively, one can view these matrices as approximations to each network's input to layer $\ell$ mapping over a discrete set of $M$ ``landmark'' points.
The representation matrices $X_\ell^A$ can be visualized as $M$ points in an $N$-dimensional space (\Cref{fig:cartoon} a).  

\textbf{Riemannian Shape Distance.}
Measuring the distance between representations $X_\ell^A$ and $X_\ell^B$ is complicated by the fact that there is no inherent correspondence between neural dimensions at layer $\ell$.
A variety of metrics based on alignment or comparing intra-network distances have been proposed (see \cite{klabunde2025similarity} for a review).
In this paper, we use the \textit{Riemannian shape distance}~\cite{Kendall1977-ty}, which alongside the closely related \textit{Procrustes distance} creates a framework for comparing deep network representations~\cite{Williams2021,ding2021grounding}.
This notion of distance considers two representations as equivalent if their point clouds (see \cref{fig:cartoon}a) can be superimposed by translation, rotation, reflections, and rescaling.
Intuitively, this would imply that the point clouds have the same geometry, and therefore carry the same linearly decodable information~\cite{pmlr-v285-harvey24a}.

To compute the Riemannian shape distance between two representation matrices $X_i \in \reals^{M \times N}$ and $X_j \in \reals^{M \times N}$, we first center the neural responses at the origin and re-scale them to unit norm:
\begin{equation}
Z_i = \frac{C X_i }{\| C X_i \|_F} \quad \text{and} \quad Z_j = \frac{C X_j }{\| C X_j \|_F}
\end{equation}
where $C = I_{M} - (1/M) 1_M 1_M^T$ is the \textit{centering matrix}.
The pre-processed matrices $Z_i$ and $Z_j$ are often called \textit{pre-shapes}.
The \emph{shape} of the representation $S_i$ is defined by the set of all rotated and reflected versions of the pre-shape: $S_i = \{ Z_i O: O \in \mathcal{O}(N) \}$,
where $O(N)$ is the orthogonal group of rotations and reflections in $N$-dimensions, which is the set of $N\times N$ matrices satisfying $O^T O = O O^T = I_N$ and $\det{O} = \pm 1$.
\emph{Shape space} $\Sigma_{N}^M$ is the space of all shapes with $M$ landmarks in $\mathbb{R}^N$.  
Then, the Riemannian shape  distance is defined as:
\begin{equation}
\label{eq:shape_dist}
\rho(X_i, X_j) = \arccos\Big{(} \sup_{O \in \mathcal{O}(N)} \text{Tr} [ Z_j^\top Z_i O  ]\Big{)}
\end{equation}
which can be interpreted as the distance along the pre-shape hypersphere between $Z_j$ and the aligned representation $Z_i^* = Z_i O^*$, where $O^*$ is the orthogonal matrix achieving the supremum in \cref{eq:shape_dist}.
\Cref{eq:shape_dist} is an intrinsic geodesic distance through shape space. 
It is measured in radians, and is the smallest angle between pre-shapes $Z_i$ and $Z_j$ on the hypersphere that can be achieved by $N$-dimensional rotations of the pre-shapes.

The shape distance, $\rho$, is symmetric and satisfies the triangle inequality.
Indeed, it is a true metric on the space of shapes,\footnote{Intuitively, a \textit{shape} is the set of equivalent \textit{pre-shapes} that can be achieved by $N$-dimensional orthogonal transformations. See \textcite{kendall2009shape,dryden2016} for further background.} 
and it defines closed-form geodesic paths between shapes as explained below.
At the same time, it embodies an intuitive minimal invariance class---representations are aligned up to rotation, reflection, translation, and global scaling---mirroring other popular methods such as linear CKA while avoiding the tuning and interpretability issues that often arise with more flexible nonlinear metrics.
Finally, unlike metrics derived from CKA, the shape space framework enables us to explicitly track how landmarks (i.e. responses to probe images) are displaced along a geodesic path, thereby enhancing human interpretability of representational change.
However, the Riemannian shape distance is related to metrics on stimulus-by-stimulus similarity matrices like $CKA$ through its equivalence to the normalized Bures similarity, described in appendix A.

\textbf{Shape geodesics and geodesic angles.} 
Given two pre-shapes $Z_i$ and $Z_j$, we can obtain a geodesic between them parametrized by $t$, $\gamma(t)$, by moving along a great circle on the pre-shape sphere.
This curve, written explicitly in appendix A, represents the shortest smooth deformation from one neural representation to another, ignoring overall orientation, scale and reflection.
%

\vspace{-1mm}
Let $\gamma_1$ and $\gamma_2$ be two geodesics between a reference pre-shape $Z_0$ and two comparison pre-shapes
$Z_1$ and $Z_2$, respectively. Their optimally aligned preshapes to initial preshape $Z_0$ are $Z_1^\ast$ and $Z_2^\ast$,
and the initial tangent vectors along $\gamma_1$ and $\gamma_2$ are known to be
\[
V_i = \frac{\rho_i}{\sin\rho_i} \left( Z_i^\ast - \cos(\rho_i) Z_0 \right),
\: \: \text{with } \rho_i = \arccos( \Tr Z_0^\top Z_i^\ast ).
\]
for $ i = 1,2.  $
These tangent vectors represent the initial direction of each geodesic path in the shape manifold.
The angle between the two shape geodesics is the angle between their tangent vectors at the reference shape:
\begin{equation}\label{eq:geo_angle}
\theta_{1,2} = \angle(\gamma_1, \gamma_2)
=
\arccos \left(
\frac{\langle V_1, V_2 \rangle}{\|V_1\|_F \cdot \|V_2\|_F}
\right).
\end{equation}
This angle will be used to quantify the similarity of two representation trajectories from the same initial shape.

\textbf{Landmark displacement.}
We can also investigate \emph{how} representation shapes are deformed as DA is applied at varying levels, 
by measuring the landmark-level displacement.
The displacement for each landmark $\Delta z_{\mu} \in \reals^N$, $\mu \in \{1, ..., M \} $,  is computed as the Euclidean displacement vector between corresponding landmarks across two optimally aligned preshapes (\Cref{fig:cartoon} a).
These displacement vectors are the rows of the matrix $\Delta Z = Z_\ell^A - Z_\ell^B O^* $, with $O^*$ being the matrix that obtains the supremum in \cref{eq:shape_dist}.
We may then study which landmarks (network inputs) are maximally or minimally distorted along a step of a trajectory in shape space, and which landmarks are distorted in similar directions.










\section{Experiments}\label{sec:experiments}

\begin{figure}[t]
    \centering    
    \includegraphics[width=\linewidth]{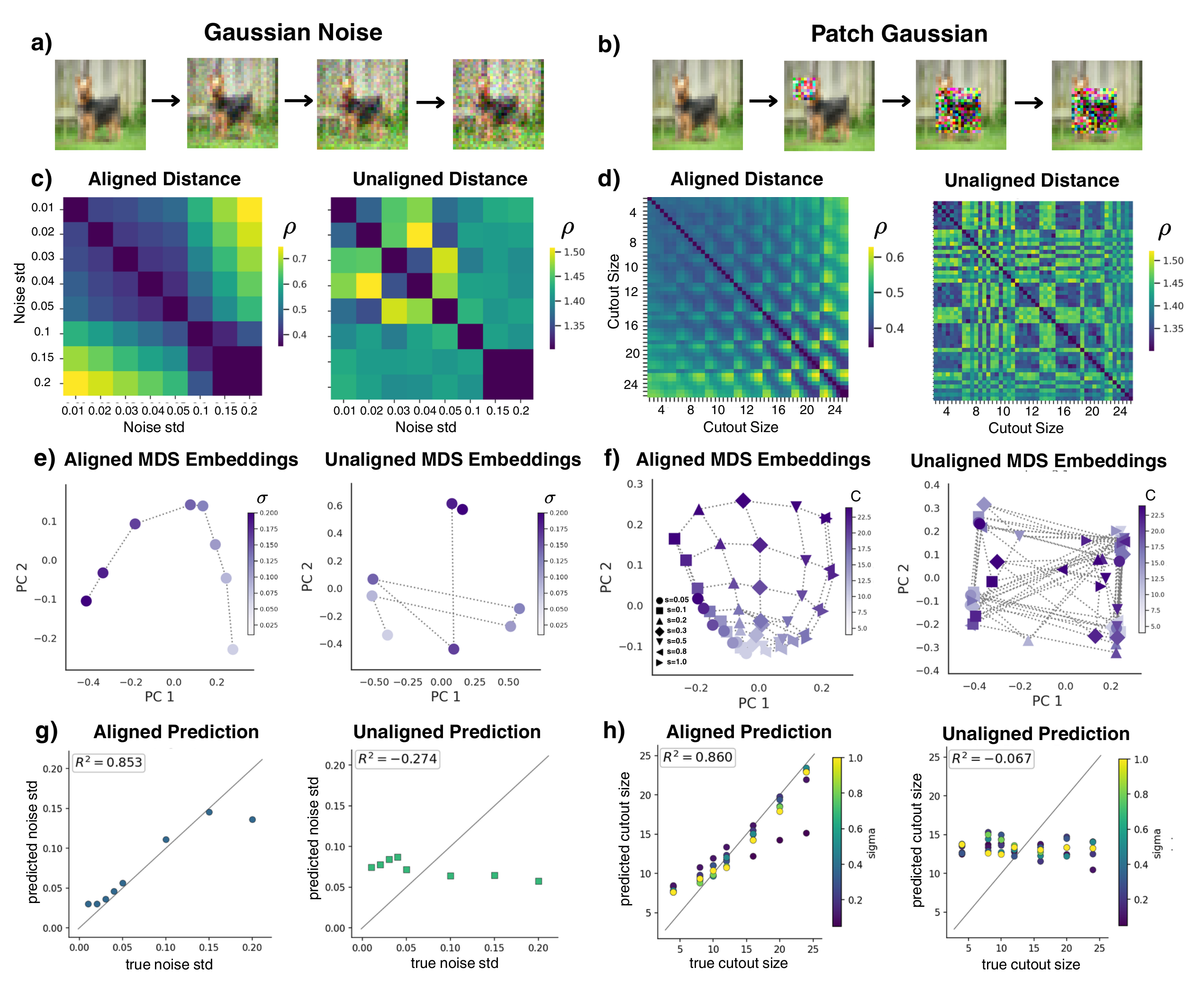} %
  \caption{  \textbf{Orthogonal alignment reveals well-behaved DA trajectories in representation shape space that are not present without alignment.} Using ResNet-18 layer 2 responses to 10,000 CIFAR-10 test set images from models trained with the same seed, pairwise representation distances show coherent organization across augmentation strengths only when the alignment step is included.
  (a, b) We compare additive Gaussian noise, parameterized by noise level $\sigma$, with patch Gaussian, parameterized by a cutout width $C$ and within-patch noise variance $s$. 
  (c, d) Distance matrices compare aligned and unaligned shape distances for Gaussian noise and patch Gaussian data augmentations at various strengths. 
  (e, f) MDS-PCA visualizations of these distances reveal smooth, ordered trajectories in shape space, and for patch Gaussian, an organized two-parameter surface over ($C,s$), which disappear when orthogonal alignment is not performed. 
  (g, h) True vs. predicted augmentation hyperparameters from leave-one-out ridge regression on aligned representation shapes of clean probe images. Points near the diagonal indicate accurate decoding of augmentation strength from shape.
  }\label{fig:fig2}
\end{figure}

To demonstrate this framework, we studied image representations formed by various vision models trained on CIFAR-10 image classification (with ImageNet \cite{deng2009imagenet} fine-tuning examples given in appendix C).
A spectrum of DA methods and magnitudes (described below) were applied during training.
After training, we collected hidden layer responses to $M = 10,000$ clean (un-augmented) probe images from the test set, and used these as the basis for shape distance calculations.  
Full training details are listed in appendix B.  

\textbf{DA steers representations along well-behaved trajectories in shape space, but not Euclidean space.}
How necessary is the alignment step, and the shape space framework generally?
One might imagine that orthogonal alignments are not necessary when comparing representations across two identical networks that are trained and initialized with the same random seed, and only differ by a small amount of DA.
To investigate this, we trained ResNet-18   \parencite{he2016deep} models with Gaussian noise data augmentation governed by noise standard deviation parameter $\sigma$, as well as with patch Gaussian data augmentation with standard deviation $s$ over a $C \times C$ square patch of pixels \parencite{lopes2019improving}.
When $C$ equals the width and height of the input images the method reduces to additive Gaussian noise (\cref{fig:fig2}a), while for large $s$, it resembles ``CutOut'' augmentation \parencite{devries2017improved} with patch size $C$ (\cref{fig:fig2}b).

Whether we restrict focus to the 1D space of Gaussian noise over the full image (\cref{fig:fig2}a, c, e, g) or the 2D space of patch Gaussian augmentations of varying patch sizes (\cref{fig:fig2}b, d, f, h) we observe consistent relational structure corresponding to DA parameters only when using the shape space framework.
This is visible in distance matrices for representations collected after layer 2 of networks trained at different augmentation levels (\cref{fig:fig2}c, d).
Each $(i,j)$ element in the distance matrix is equal to the shape distance between representations $i$ and $j$ trained with different DA hyperparameters, with (left) and without (right) the alignment step in the definition \cref{eq:shape_dist}. 
To visualize this relational structure, we use the MDS-PCA procedure of \parencite{Williams2021}, using multidimensional scaling to find points in a 200 dimensional Euclidean space with pairwise distances that well-approximate the shape distances, and plotting the leading PCA axes. 
\Cref{fig:fig2} e, f show for both DA methods the destruction of shape organization with DA hyperparameters when orthogonal alignment is not performed.
\Cref{fig:cartoon} and \Cref{fig:rot_sheer} replicate this experiment using augmentation pairs CutOut with random crop, and rotation with sheer, and different geometric patterns can be observed for DA methods.
\Cref{fig:wideresnet} replicates this experiment with pretrained ResNet18 \cite{zagoruyko2016wide} fine-tuned using the ImageNet dataset \cite{deng2009imagenet}.

To quantitatively measure how regularly augmentation steers representations through shape space, 
we investigated how well the DA hyperparameter used during training for a held-out representation could be predicted linearly using the representations of the clean probe inputs alone.  
 We first aligned each representation shape to the un-augmented shape by the optimal orthogonal transformation and then flattened the aligned shapes into vectors. 
 We fit a ridge regressor to these vectors to predict the DA hyperparameter used to train each model, and evaluated prediction with leave-one-shape-out cross-validation, fitting on the remaining aligned shapes, and predicting the held-out hyperparameter value. 
Geometrically, this can be viewed as a first-order approximation to tangent-space regression at the un-augmented representation shape. 
We find that the Gaussian noise and patch size hyperparameters for held-out representations can be well-predicted linearly from representations trained with those augmentations (\cref{fig:fig2} g, h), indicating that representations are not scattered arbitrarily in shape space; they are organized in a systematic way with respect to the DA parameters, but only when the orthogonal alignment step is first performed.

\begin{figure}[t!]
    \centering    
    \includegraphics[width=\linewidth]{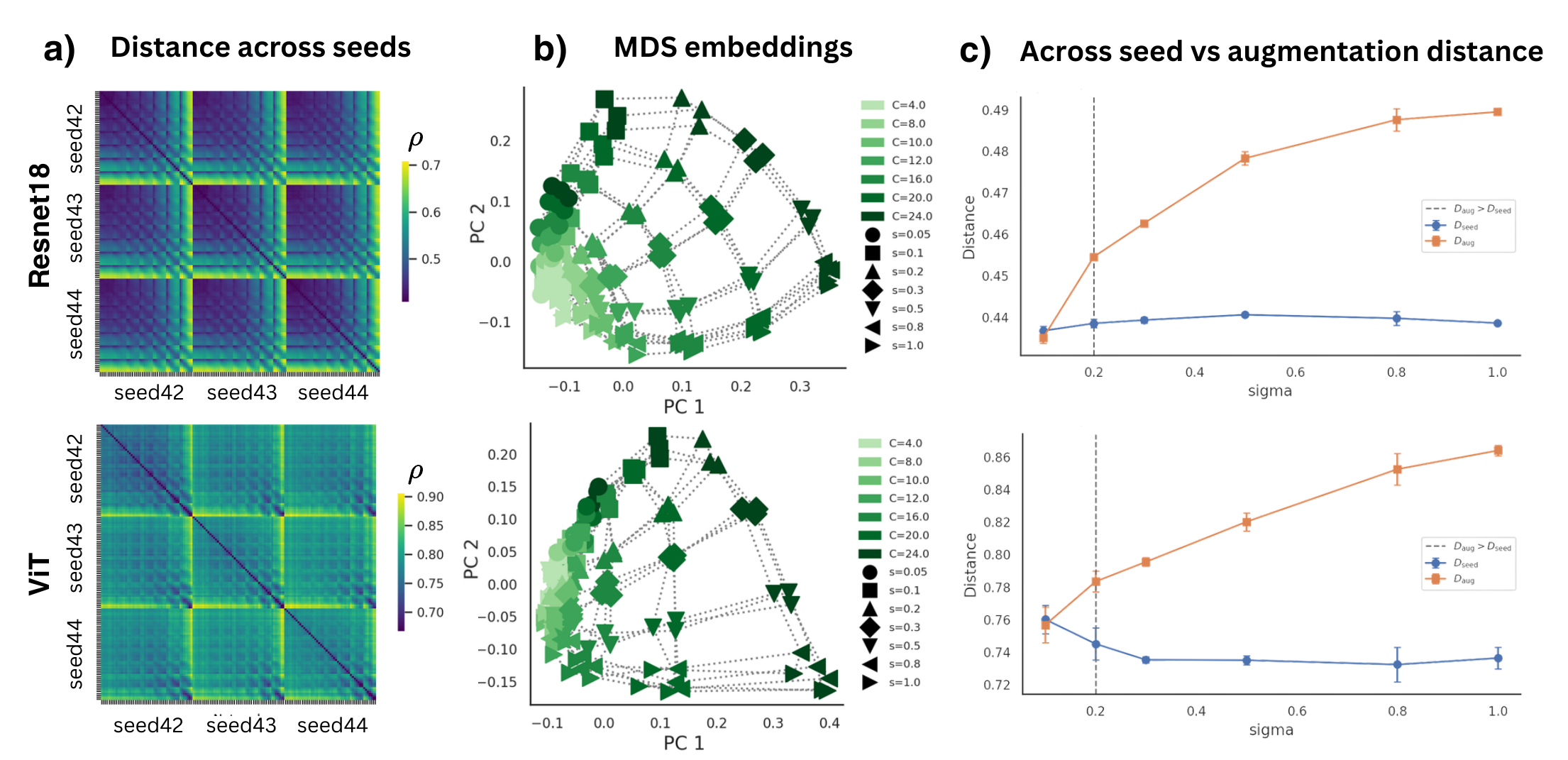} %
  \caption{\textbf{Augmentation-driven representation changes in shape space can be compared directly with seed-to-seed variability.} Patch Gaussian DA is applied during training of ResNet-18 (top) and ViT (bottom) for a grid of ($C,s$) values. (a) Cross-seed distance matrices show repeated within-seed structure, (b) MDS-PCA suggests similar augmentation trajectories across three random seeds.  (c) $D_{\text{aug}}$ and $D_{\text{seed}}$ averaged over three pairs of random seeds, with error bars corresponding to the standard deviations, as Gaussian noise DA strength $s$ is increased at fixed patch size of $C=12$ (top plot) and $C=24$ (bottom plot). Divergence of these two curves indicates when changes due to the Gaussian noise parameter exceed those due to random seed. }\label{fig:random_seeds}
\end{figure}

\textbf{The magnitude of representational shape change due to random seed gives a natural scale to evaluate the size of changes due to DA.}
How sensitive are representations to small changes in DA, when compared with the effect of changing the random seed used for initialization?
In light of recently observed ``butterfly effects" in which small perturbations early in training lead to training trajectories that drive models into distinct loss minima \parencite{altntas2025}, we are interested in understanding how the effect of random initialization compares with small changes in DA hyperparameter.  
The shape metric space framework gives us a natural way to compare the scales of these two sources of variability in terms of the Riemannian shape distance between their learned representations.  
The distance matrices in \cref{fig:random_seeds} (a) show the Riemannian shape distance between learned representations from the penultimate layer of ResNet-18 and ViT  trained with DA method patch Gaussian over a range of ($C,s$) values, for three different random seeds 42, 43, 44.
Each distance matrix shows repeated off-diagonal blocks corresponding to the three random seeds, suggesting that the representations share similar within-seed structure as the Gaussian noise variance and patch size are varied. 
This can be visualized in the MDS-PCA plots in \cref{fig:random_seeds} (b), where trajectories corresponding to changing the DA hyperparameter values at a fixed random seed are shown connected by dotted lines. 
From the MDS-PCA visualizations, it appears that there is a transition point between a regime in which the scale of representational change due to these data augmentation methods is similar to that induced by changing the random seed (small $C$, small $s$ regime), to a regime where the scale of change due to the data augmentation is larger than the changes due to random seed.  
To quantify this, we define two measured distance scales:
\vspace{-1mm}
\begin{equation}
    D_{\text{aug}} =  \frac{1}{2}(\rho(X_0^{(i)},X_p^{(i)} ) + \rho(X_0^{(j)},X_p^{(j)} ) ), \hspace{5mm} D_{\text{seed}}=  \frac{1}{2} (\rho(X_0^{(i)},X_0^{(j)}) + \rho(X_p^{(i)},X_p^{(j)} ))
\end{equation}
where $X_0^{(i)}$ is a representation matrix learned without DA from random seed $i$, and $X_p^{(i)}$ is a representation matrix learned with a DA hyperparameter $p$.
If $D_{\text{aug}} > D_{\text{seed}} $ , this indicates that the scale of representation shape changes due to the change in data augmentation hyperparameter tends to be larger than changes due to random seed.  
As the data augmentation hyperparameter approaches $0$, $D_{\text{aug}} = D_{\text{seed}} $.
When these two distance scales diverge from one another as a hyperparameter $p$ is increased is a measure of how large of a DA parameter change
is required before the resulting magnitude of representation shape change is distinguishable from a shape change due to random seed.  
In \Cref{fig:random_seeds} (c) we compute $D_{\text{aug}}$ and $D_{\text{seed}} $ averaged over three pairs of random seeds for both ResNet-18 and ViT as $s$ is varied at constant $C$.
We find that these distance scales indeed separate around $s = 0.2$, providing a sense of ``natural units" for this hyperparameter.

\begin{figure}[t]
    \centering    
    \includegraphics[width=\linewidth]{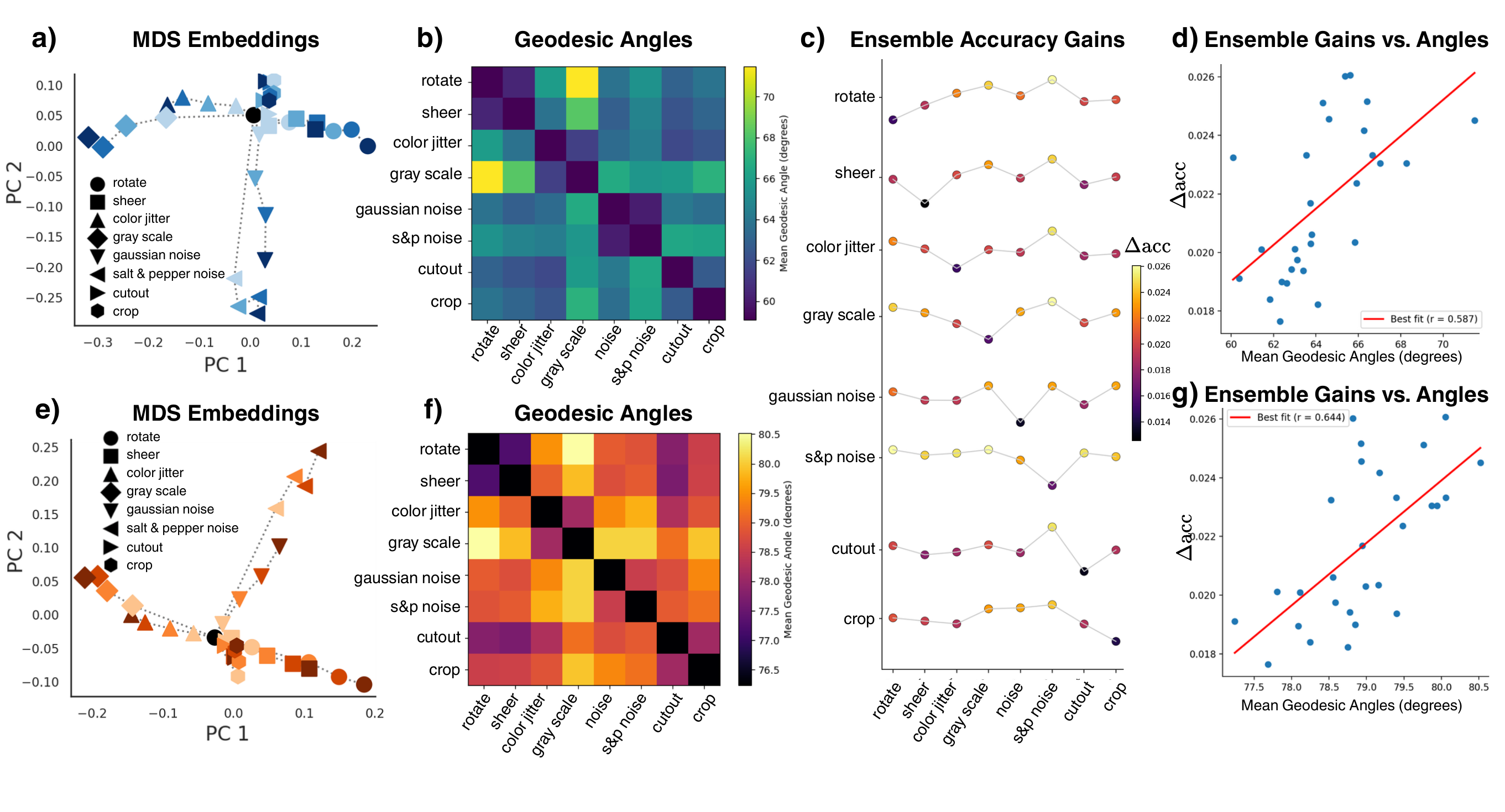} %
  \caption{ \textbf{Different augmentation types produce distinct representational changes, whose diversity predicts ensemble gains.} 
(a) MDS-PCA embedding of representation shapes for ResNet18 layer 2.1 trained with 8 distinct augmentation types applied individually. Each trajectory connected with dotted lines corresponds to a single augmentation method applied at different magnitudes listed in Table \ref{tab:augmentation}.  Increasing augmentation strength moves the model representations away from the representation shape measured after training with no data augmentation (black circle).
(b) Heatmap of pairwise geodesic angles between the augmentation trajectories, measured with the un-augmented model as the vertex as in \Cref{fig:cartoon} 
(c) Improvement in test accuracy of ensembled models over average accuracy of model ensemble.  Models trained separately with different augmentation methods, then ensembled pairwise across methods. 
(d) Scatter plot showing the correlation between geodesic angle and the improvement in ensemble accuracy across pairs of augmentations, suggesting that geometric diversity is related with larger ensemble gains. (e,f,g) Analogous MDS-PCA embedding and geodesic angle measurements for AvgPool layer corresponding to panels (a,b,d).}\label{fig:single_augments}
\end{figure}

\textbf{Different data augmentation methods move representations in different directions in shape space, and the angle between trajectories is predictive of model ensembling gains.}
We now study the effect of classic DA methods on representation shape geometry when applied individually.  
The embedding of representations into shape space established in \Cref{sec:methods} allows us to study geometric properties of representation \emph{trajectories} through shape space, as a function of DA hyperparameter.
\Cref{fig:single_augments} (a,e) show MDS-PCA visualizations of these trajectories for eight different DA methods in ResNet-18, each applied at four different magnitudes. 
Dotted lines connect points of sequential augmentation hyperparameter magnitude listed in appendix B table \ref{tab:augmentation}, starting from the black point corresponding with no DA.
(a) shows trajectories of representation shapes collected after layer 2, while (e) repeats the same analysis on the penultimate avgpool layer.
The mean geodesic angle for two trajectories is computed as the angle in the tangent space of the un-augmented shape formed by geodesics between the un-augmented shape and steps along two DA trajectories (\Cref{eq:geo_angle}), averaged over all steps in the trajectories. 
\Cref{fig:single_augments} (b) and (f) show heatmaps of the mean geodesic angles measured between all DA trajectories for layer 2 (b) and avgpool (f). 
Some clustering by DA type (geometric, photometric, noise, occlusion) is observed in both the MDS-PCA trajectory visualizations and the geodesic angles.  

It is important to note that the dimensionality of shape space is known \parencite{dryden2016} to be $N(M - 1) - 1 - N(N-1)/2 \approx 10^6$ for our deep neural network representations, even after dimensionality reduction was performed on the activations (appendix B). 
In this high dimensional space, two randomly chosen directions in the tangent space of a particular shape are highly likely to have an angle very close to \(90^{\circ }\) and observations by chance of significantly acute angles are overwhelmingly rare \parencite{vershynin2018high}, 
suggesting that the measured angles between $\sim 60^\circ$ and $\sim 81^\circ$ could indicate non-trivial similarity between trajectory types.
As a practical test of this idea, we measure the test accuracy of ensembles of pairs of DA methods. Each ensemble consists of two sets of trained models, $\mathcal{M}_A, \mathcal{M}_B$, from each DA trajectory in \Cref{fig:single_augments} a .  
Models are combined using soft voting by averaging their predicted class probabilities: 
$p_{\mathrm{ens}}(y \mid x; A,B)
=
\frac{1}{|\mathcal{M}_A| + |\mathcal{M}_B|}
\sum_{f \in \mathcal{M}_A \cup \mathcal{M}_B}
p_f(y \mid x)$, where $p_f(y \mid x) = \operatorname{softmax}(z_f(x))_y$ for model $f$. 
We quantify the benefit of ensembling using the ensembled classification accuracy improvement over the average constituent model accuracy,
$
\Delta \mathrm{Acc}(A,B)
=
\mathrm{Acc}_{\mathrm{ens}}(A,B)
-
\frac{1}{|\mathcal{M}_A| + |\mathcal{M}_B|}
\sum_{f \in \mathcal{M}_A \cup \mathcal{M}_B}
\mathrm{Acc}(f).
$ 
We find that ensemble gains of DA pairs correlate with the mean geodesic angle between their trajectories (\Cref{fig:single_augments}), suggesting that improvements from ensembling are connected with the geometric diversity of the learned representations of the individual models.  
%
\begin{figure}[t!]
    \centering    
    \includegraphics[width=1.0\linewidth]{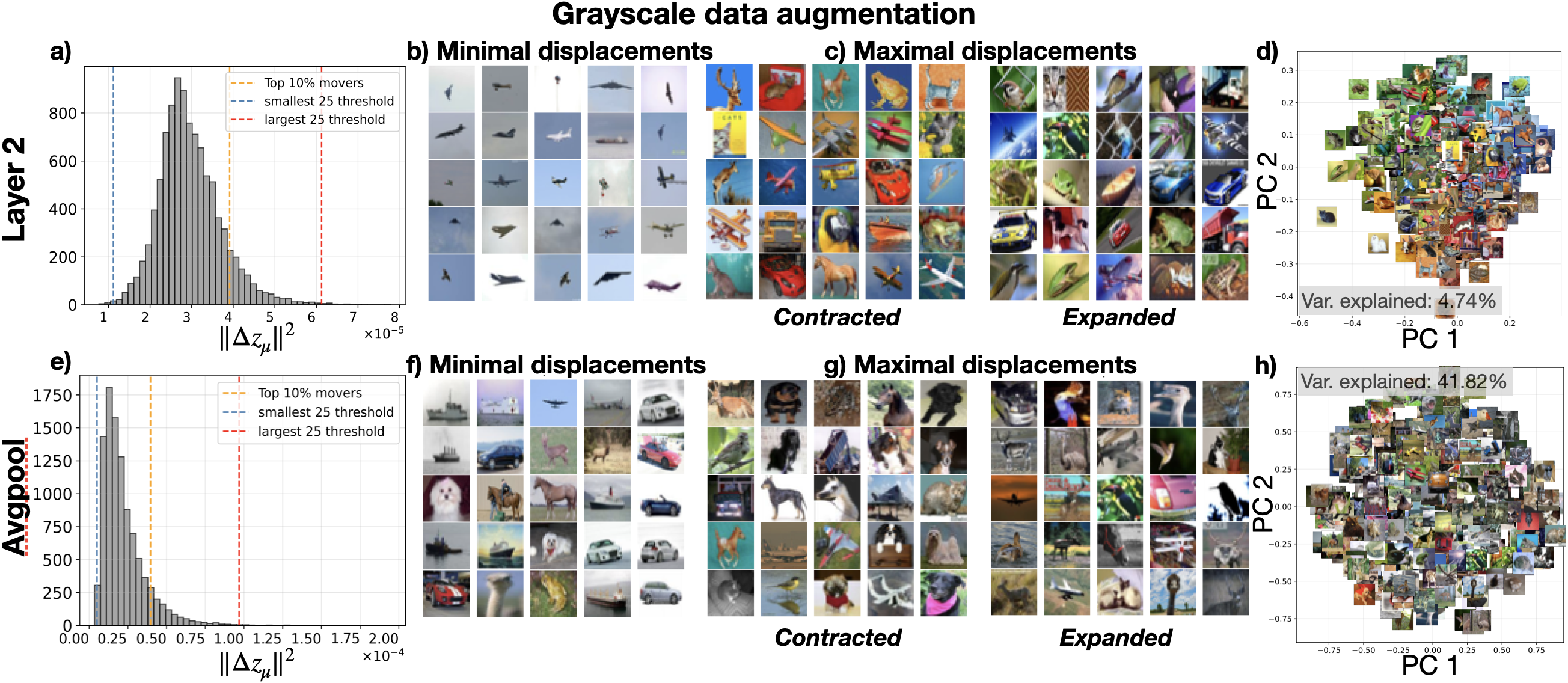} %
\caption{\textbf{DA produces layer-dependent landmark distortions.} 
Landmark displacement analysis is performed comparing ResNet-18 representations trained without DA with those trained with grayscale DA for layer 2 (top) and avgpool (bottom). (a,e) Histograms of landmark displacement magnitudes across probe images; (b,f) 25 minimally displaced landmarks; (c,g) maximally displaced landmarks split into those contracting toward the origin and expanding away from the origin; (d,h) first two principal components of the row-normalized displacement matrix $\Delta Z$.}
\label{fig:maxminimages}
\end{figure}

\textbf{Representation shape changes due to DA can be inspected with landmark displacements, which show layer-dependent structure.}
Over a step along a representation shape trajectory, each landmark (probe input) $\mu = 1, ... ,M$ is displaced by $\Delta z_\mu \in \reals^N$ (\Cref{fig:cartoon}), whose magnitudes and directions encode which aspects of the representation is distorted after rotations, reflections and scalings are removed.
This serves as a window into the layer-by-layer mechanism by which the DA process reshapes learned representations.  

\Cref{fig:maxminimages} studies properties of the landmark displacements for ResNet-18, comparing an un-augmented representation to those resulting from training from the same random seed but with a small amount of ``grayscale" DA (3\% chance of image grayscaling during training).
Panels on the top row correspond with representations after layer 2, while the bottom row show the same analysis for the penultimate avgpool representations.
The distributions of displacement magnitudes shows that in both layers a small number of landmarks are moved significantly more or less than the average displacement (\Cref{fig:maxminimages} a, e).  
Image tiles in \Cref{fig:maxminimages} (b, c, f, and g) (and see also \Cref{fig:sheer}, \Cref{fig:grayscale} and \Cref{fig:gaussian})
show the maximally and minimally moved landmarks from the distributions in panels (a) and (e), with the maximally displaced landmarks divided into landmarks that move toward the origin (``contracted" images) and those that move away from the origin (``expanded" images).  
These expanding and contracting landmarks are of potential interest because they indicate sets of images whose features at this stage of the model are attenuated or amplified by the effect of the particular data augmentation during training.  
\Cref{fig:maxminimages} (d) and (h) show the top two principal components of the row-normalized landmark displacement matrix $\Delta Z$, as a way of probing whether groups of landmarks tend to be displaced in similar directions.

From \cref{fig:maxminimages} we can see that in the avgpool layer, while the landmark displacements are lower dimensional in terms of variance explained by the top two principal components (h), patterns in the maximally or minimally displaced landmarks or their directions are not obvious.
However, in layer 2, 
the most affected landmarks seem related to each other in terms of lower-level image features such as bright background colors and diagonal edges.
This, together with additional examples for different augmentation methods presented in appendix C, suggests that the effect of DA on representation is both layer and DA method dependent.
We conjecture that the properties and patterns of the landmark distortions could help elucidate the mechanisms by which DA or other hyperparameters deform representation shape layer by layer, and this would be an exciting avenue for future work.



\section{Discussion}

\textbf{Limitations.} 
This work is limited to image classification, classical DA methods, and post-hoc analysis of a fixed sample of unaltered probe images from the test set; extending the approach to other modalities, larger models, learned augmentation policies, and training dynamics is left to future work. 
The Riemannian shape distance is also expensive for high-dimensional convolutional representations, so dimensionality reduction of the representation matrices to a manageable dimensionality while preserving a high amount of variance is often required in practice (see appendix B). 

Beyond the specific case of DA, this work suggests broader uses for shape space analysis as a general tool for studying how training choices steer learned functions. 
This framework offers a common language for comparing models trained with different optimizers, regularizers, architectures, supervision, or fine-tuning protocols, using tools from Riemannian geometry. 
Rather than asking only whether two settings achieve similar accuracy, one can ask whether they move representations along similar paths, or whether small hyperparameter adjustments trigger abrupt transitions into new representational regimes. 
In this sense, shape space geometry could become useful for hyperparameter search, augmentation policy design, principled ensemble construction, or understanding representational changes due to fine-tuning, especially in settings where standard validation metrics are too coarse to distinguish between many near-equivalent training choices.

\printbibliography

@inproceedings{deng2009imagenet,
  title={ImageNet: A large-scale hierarchical image database},
  author={Deng, Jia and Dong, Wei and Socher, Richard and Li, Li-Jia and Li, Kai and Fei-Fei, Li},
  booktitle={2009 IEEE Conference on Computer Vision and Pattern Recognition},
  pages={248--255},
  year={2009},
  organization={IEEE}
}

@article{zagoruyko2016wide,
  title={Wide residual networks},
  author={Zagoruyko, Sergey and Komodakis, Nikos},
  journal={arXiv preprint arXiv:1605.07146},
  year={2016}
}

@article{klabunde2025similarity,
  title={Similarity of neural network models: A survey of functional and representational measures},
  author={Klabunde, Max and Schumacher, Tobias and Strohmaier, Markus and Lemmerich, Florian},
  journal={ACM Computing Surveys},
  volume={57},
  number={9},
  pages={1--52},
  year={2025},
  publisher={ACM New York, NY}
}

@article{nguyen2020,
  author       = {Thao Nguyen and
                  Maithra Raghu and
                  Simon Kornblith},
  title        = {Do Wide and Deep Networks Learn the Same Things? Uncovering How Neural
                  Network Representations Vary with Width and Depth},
  journal      = {CoRR},
  volume       = {abs/2010.15327},
  year         = {2020},
  url          = {https://arxiv.org/abs/2010.15327},
  eprinttype    = {arXiv},
  eprint       = {2010.15327},
  bibsource    = {dblp computer science bibliography, https://dblp.org}
}

@article{devries2017improved,
  title={Improved regularization of convolutional neural networks with cutout},
  author={DeVries, Terrance and Taylor, Graham W},
  journal={arXiv preprint arXiv:1708.04552},
  year={2017}
}

@InProceedings{pmlr-v285-harvey24a,
  title = 	 {What Representational Similarity Measures Imply about Decodable Information},
  author =       {Harvey, Sarah E and Lipshutz, David and Williams, Alex H},
  booktitle = 	 {Proceedings of UniReps: the Second Edition of the Workshop on Unifying Representations in Neural Models},
  pages = 	 {140--151},
  year = 	 {2024},
  editor = 	 {Fumero, Marco and Domine, Clementine and Lähner, Zorah and Crisostomi, Donato and Moschella, Luca and Stachenfeld, Kimberly},
  volume = 	 {285},
  series = 	 {Proceedings of Machine Learning Research},
  month = 	 {14 Dec},
  publisher =    {PMLR},
}

@book{kendall2009shape,
  title={Shape and shape theory},
  author={Kendall, David George and Barden, Dennis and Carne, Thomas K and Le, Huiling},
  year={2009},
  publisher={John Wiley \& Sons}
}

@article{ding2021grounding,
  title={Grounding representation similarity through statistical testing},
  author={Ding, Frances and Denain, Jean-Stanislas and Steinhardt, Jacob},
  journal={Advances in Neural Information Processing Systems},
  volume={34},
  pages={1556--1568},
  year={2021}
}

@article{lopes2019improving,
  title={Improving robustness without sacrificing accuracy with patch gaussian augmentation},
  author={Lopes, Raphael Gontijo and Yin, Dong and Poole, Ben and Gilmer, Justin and Cubuk, Ekin D},
  journal={arXiv preprint arXiv:1906.02611},
  year={2019}
}

@book{vershynin2018high,
  title     = {High-Dimensional Probability: An Introduction with Applications in Data Science},
  author    = {Vershynin, Roman},
  year      = {2018},
  publisher = {Cambridge University Press},
  series    = {Cambridge Series in Statistical and Probabilistic Mathematics},
  isbn      = {9781108415194}
}

@book{dryden2016,
  title     = {Statistical Shape Analysis: With Applications in R},
  author    = {Dryden, Ian L. and Mardia, Kanti V.},
  year      = {2016},
  edition   = {2nd},
  publisher = {John Wiley \& Sons},
  isbn      = {978-0-470-69962-1}
}

@article{smith2018,
  title        = {A disciplined approach to neural network hyper-parameters: Part 1 -- learning rate, batch size, momentum, and weight decay},
  author       = {Smith, Leslie N.},
  journal      = {CoRR},
  volume       = {abs/1803.09820},
  year         = {2018},
  url          = {https://arxiv.org/abs/1803.09820}
}

@article {Olshausen4700,
	author = {Olshausen, BA and Anderson, CH and Van Essen, DC},
	title = {A neurobiological model of visual attention and invariant pattern recognition based on dynamic routing of information},
	volume = {13},
	number = {11},
	pages = {4700--4719},
	year = {1993},
	doi = {10.1523/JNEUROSCI.13-11-04700.1993},
	publisher = {Society for Neuroscience},
	issn = {0270-6474},
	URL = {https://www.jneurosci.org/content/13/11/4700},
	eprint = {https://www.jneurosci.org/content/13/11/4700.full.pdf},
	journal = {Journal of Neuroscience}
}

@article{riesenhuber1999hierarchical,
  title={Hierarchical models of object recognition in cortex},
  author={Riesenhuber, Maximilian and Poggio, Tomaso},
  journal={Nature neuroscience},
  volume={2},
  number={11},
  pages={1019--1025},
  year={1999},
  publisher={Nature Publishing Group}
}

@article{dicarlo2007untangling,
  title={Untangling invariant object recognition},
  author={DiCarlo, James J and Cox, David D},
  journal={Trends in cognitive sciences},
  volume={11},
  number={8},
  pages={333--341},
  year={2007},
  publisher={Elsevier}
}

@InProceedings{Lenc_2015_CVPR,
author = {Lenc, Karel and Vedaldi, Andrea},
title = {Understanding Image Representations by Measuring Their Equivariance and Equivalence},
booktitle = {Proceedings of the IEEE Conference on Computer Vision and Pattern Recognition (CVPR)},
month = {June},
year = {2015}
}

@article{chen2020,
  title={A group-theoretic framework for data augmentation},
  author={Chen, Shuxiao and Dobriban, Edgar and Lee, Jane H},
  journal={Journal of Machine Learning Research},
  volume={21},
  number={245},
  pages={1--71},
  year={2020}
}

@article{bishop1995,
  title={Training with noise is equivalent to Tikhonov regularization},
  author={Bishop, Christopher M.},
  journal={Neural Computation},
  volume={7},
  number={1},
  pages={108--116},
  year={1995},
  publisher={MIT Press},
  doi={10.1162/neco.1995.7.1.108},
  url={https://doi.org/10.1162/neco.1995.7.1.108}
}

@inproceedings{Williams2021,
 author = {Williams, Alex H and Kunz, Erin and Kornblith, Simon and Linderman, Scott},
 booktitle = {Advances in Neural Information Processing Systems},
 editor = {M. Ranzato and A. Beygelzimer and Y. Dauphin and P.S. Liang and J. Wortman Vaughan},
 pages = {4738--4750},
 publisher = {Curran Associates, Inc.},
 title = {Generalized Shape Metrics on Neural Representations},
 volume = {34},
 year = {2021}
}

@InProceedings{Lange2023,
  title = 	 {Deep Networks as Paths on the Manifold of Neural Representations},
  author =       {Lange, Richard D and Kwok, Devin and Matelsky, Jordan Kyle and Wang, Xinyue and Rolnick, David and Kording, Konrad},
  booktitle = 	 {Proceedings of 2nd Annual Workshop on Topology, Algebra, and Geometry in Machine Learning (TAG-ML)},
  pages = 	 {102--133},
  year = 	 {2023},
  editor = 	 {Doster, Timothy and Emerson, Tegan and Kvinge, Henry and Miolane, Nina and Papillon, Mathilde and Rieck, Bastian and Sanborn, Sophia},
  volume = 	 {221},
  series = 	 {Proceedings of Machine Learning Research},
  month = 	 {28 Jul},
  publisher =    {PMLR},
  url = 	 {https://proceedings.mlr.press/v221/lange23a.html},
}

@ARTICLE{Kendall1977-ty,
  title     = "The Diffusion of Shape",
  author    = "Kendall, D G",
  journal   = "Adv. Appl. Probab.",
  publisher = "Applied Probability Trust",
  volume    =  9,
  number    =  3,
  pages     = "428--430",
  year      =  1977
}

@InProceedings{harvey24a,
  title = 	 {Duality of Bures and Shape Distances with Implications for Comparing Neural Representations},
  author =       {Harvey, Sarah E. and Larsen, Brett W. and Williams, Alex H.},
  booktitle = 	 {Proceedings of UniReps: the First Workshop on Unifying Representations in Neural Models},
  pages = 	 {11--26},
  year = 	 {2024},
  editor = 	 {Fumero, Marco and Rodolá, Emanuele and Domine, Clementine and Locatello, Francesco and Dziugaite, Karolina and Mathilde, Caron},
  volume = 	 {243},
  series = 	 {Proceedings of Machine Learning Research},
  month = 	 {15 Dec},
  publisher =    {PMLR},
  url = 	 {https://proceedings.mlr.press/v243/harvey24a.html}
}

@misc{dao2019,
      title={A Kernel Theory of Modern Data Augmentation}, 
      author={Tri Dao and Albert Gu and Alexander J. Ratner and Virginia Smith and Christopher De Sa and Christopher Ré},
      year={2019},
      eprint={1803.06084},
      archivePrefix={arXiv},
      primaryClass={cs.LG},
      url={https://arxiv.org/abs/1803.06084}, 
}

@misc{kornblith2019,
      title={Similarity of Neural Network Representations Revisited}, 
      author={Simon Kornblith and Mohammad Norouzi and Honglak Lee and Geoffrey Hinton},
      year={2019},
      eprint={1905.00414},
      archivePrefix={arXiv},
      primaryClass={cs.LG},
      url={https://arxiv.org/abs/1905.00414}, 
}

@article{shorten2019,
author = {Shorten, Connor and Khoshgoftaar, Taghi},
year = {2019},
month = {07},
pages = {},
title = {A survey on Image Data Augmentation for Deep Learning},
volume = {6},
journal = {Journal of Big Data},
doi = {10.1186/s40537-019-0197-0}
}

@inproceedings{altntas2025,
title={The Butterfly Effect: Neural Network Training Trajectories Are Highly Sensitive to Initial Conditions},
author={G{\"u}l Sena Alt{\i}nta{\c{s}} and Devin Kwok and Colin Raffel and David Rolnick},
booktitle={Forty-second International Conference on Machine Learning},
year={2025},
url={https://openreview.net/forum?id=L1Bm396P0X}
}

@misc{wang2025,
      title={A Comprehensive Survey on Data Augmentation}, 
      author={Zaitian Wang and Pengfei Wang and Kunpeng Liu and Pengyang Wang and Yanjie Fu and Chang-Tien Lu and Charu C. Aggarwal and Jian Pei and Yuanchun Zhou},
      year={2025},
      eprint={2405.09591},
      archivePrefix={arXiv},
      primaryClass={cs.LG},
      url={https://arxiv.org/abs/2405.09591}, 
}

@inproceedings{he2016deep,
  title={Deep residual learning for image recognition},
  author={He, Kaiming and Zhang, Xiangyu and Ren, Shaoqing and Sun, Jian},
  booktitle={Proceedings of the IEEE conference on computer vision and pattern recognition},
  pages={770--778},
  year={2016}
}

\newpage
\appendix

\section{Riemannian shape distance details}\label{app: appendixA}

The solution to the optimization problem in \Cref{eq:shape_dist} is available in closed form as
\begin{equation}\label{eq:dist_closedform}
    \rho(X_i, X_j) = \arccos\Big{(} \sum_{k = 1}^N \sigma_k \Big{)}
\end{equation}
where $Z_j^\top Z_i = U\Sigma V^\top$ is the singular value decomposition of $Z_j^\top Z_i$ and $\sigma_k$ are the diagonal elements of $\Sigma$.

\textbf{Shape geodesics.}

Given two pre-shapes $Z_i$ and $Z_j$, we can obtain a geodesic between them by moving along a great circle on the pre-shape sphere.
Concretely, if $O^*$ is the orthogonal transformation achieving the supremum in \cref{eq:shape_dist}, then for any value of $t \in [0, 1]$,
$$
\gamma(t)
=
\frac{\sin((1 - t)\rho)}{\sin\rho}\, Z_j
+
\frac{\sin(t \rho)}{\sin\rho}\, Z_i O^\ast
$$
defines a new pre-shape along the geodesic curve that represents the shortest smooth deformation from one neural representation to another, ignoring overall orientation, scale and reflection.
Note that $\gamma(t)$ represents another \textit{pre-shape}, which can be viewed as a exemplar for the representation's shape.

\textbf{Connection to fidelity and normalized Bures similarity.}

The Riemannian shape distance also admits an equivalent formulation in terms of the
fidelity between the corresponding stimulus-by-stimulus kernel matrices. Let
\[
K_i = C X_i X_i^\top C,
\qquad
K_j = C X_j X_j^\top C,
\]
where $C = I_M - \frac{1}{M}\mathbf{1}\mathbf{1}^\top$ is the centering matrix used in
Eq.~(1). Define the fidelity between positive semidefinite matrices by
\[
F(A,B) \;=\; \mathrm{Tr}\!\left[\left(A^{1/2} B A^{1/2}\right)^{1/2}\right].
\]
Then the normalized Bures similarity between $K_i$ and $K_j$ is
\[
\mathrm{NBS}(K_i,K_j)
\;=\;
\frac{F(K_i,K_j)}{\sqrt{\mathrm{Tr}(K_i)\,\mathrm{Tr}(K_j)}},
\]
and Harvey et al.~\cite{harvey24a} show that this quantity is exactly the cosine of the
Riemannian shape distance:
\[
\mathrm{NBS}(K_i,K_j) \;=\; \cos\!\bigl(\rho(X_i,X_j)\bigr).
\]
Equivalently,
\[
\rho(X_i,X_j)
\;=\;
\arccos\!\left(
\frac{F(K_i,K_j)}{\sqrt{\mathrm{Tr}(K_i)\,\mathrm{Tr}(K_j)}}
\right).
\]

Thus, our shape-space metric can also be interpreted as the angular form of a
fidelity-based comparison between the centered linear kernel matrices induced by the two
representations. In the special case where $X_i$ and $X_j$ are already centered across
probe inputs, one may simply write $K_i = X_i X_i^\top$ and $K_j = X_j X_j^\top$.

\section{Training details}\label{app: appendixB}

We train all our models on CIFAR-10 with four different architectures using 4 NVIDIA A100 GPUs and the standard train/test split. Unless otherwise specified, training hyperparameters are fixed for each architecture and are reported in Table \ref{tab:hyperparams}. Hyperparameters are tuned such that they reach a reasonable classification performance while also maintaining a smooth and stable learning curve. 

Activations produced by convolutional layers of size $M \times C\times H\times W$, where 
$C$ is the number of channels and $H\times W$ are the spatial dimensions, were flattened into representation matrices $X_i \in \reals^{M\times CHW}$ (as is common, see e.g. \cite{kornblith2019}). 

Representations formed by hidden layers are often extremely large in terms of the number of activations $N=CHW$, and the decomposition required to compute \Cref{eq:dist_closedform} can become computationally expensive.
If $N > 1000$, dimensionality of the representations was reduced by discarding all but the top $1000$ principal components, with the observation that in all representations studied this projection preserved at least $ 75\%$ of the variance.   

\renewcommand{\thefigure}{A-\arabic{figure}}
\setcounter{figure}{0}

\begin{table}[h]
\centering
\caption{Training Hyperparameters for CIFAR-10 Models}
\label{tab:hyperparams}
\begin{tabular}{@{}lcccc@{}}
\toprule
\textbf{Hyperparameter} 
& \textbf{ResNet-18} 
& \textbf{VGG-16} 
& \textbf{DenseNet-40} 
& \textbf{ViT-Tiny} \\ 
\midrule
Optimizer & SGD & SGD & SGD & AdamW \\
Momentum & 0.9 & 0.9 & 0.9 & -- \\
Weight Decay & $5\times10^{-4}$ & $5\times10^{-4}$ & 0.01 & $5\times10^{-2}$ \\
Batch Size & 256 & 256 & 256 & 256 \\
Epochs & 100 & 100 & 100 & 50 \\
Learning Rate Schedule & One-cycle & One-cycle & One-cycle & One-cycle \\
Max Learning Rate & 0.03 & 0.05 & 0.01 & 0.001 \\
Warmup Epochs & 0 & 0 & 5 & 5 \\
\bottomrule
\end{tabular}
\end{table}

The data augmentations used in our experiments for \Cref{fig:single_augments} are listed in Table \ref{tab:augmentation}. For each experiment, we fix the augmentation type and sweep over a predefined set of magnitudes, while keeping the model architecture, optimizer, learning rate schedule, and number of epochs unchanged. All augmentations are applied stochastically on-the-fly during training.

\begin{table}[h]
\centering
\caption{Data Augmentation Hyperparameters}
\label{tab:augmentation}
\begin{tabular}{@{}ll@{}}
\toprule
\textbf{Augmentation} & \textbf{Value(s)} \\ 
\midrule
Rotation & $\pm \theta \in \{4, 12, 18, 24\}$ degrees \\
Shear & $\pm s \in \{4, 12, 18, 24\}$ degrees \\
Color Jitter & $c \in \{0.1, 0.2, 0.3, 0.4\}$ for Brightness/Contrast/Saturation, hue $= c/4$ \\
Grayscale & Convert images to grayscale with $p\in \{0.03, 0.05, 0.08, 0.1\}$ \\
Gaussian Noise & $\sigma \in \{ 0.01 ,0.02, 0.03,  0.04\}$ \\
Salt-and-Pepper Noise & Pixel probability $p \in \{ 0.05,0.1,0.15,0.2 \}$ \\
Random Crop & Scale $c \in \{0.6, 0.7, 0.8, 0.9\}$ \\
Cutout & size k $\in \{2,4,6,8\} $\\
\bottomrule
\end{tabular}
\end{table}

\newpage

\section{Supplementary figures}\label{sec:appendixC}

\begin{figure}[h]
    \centering    
    \includegraphics[width=\linewidth]{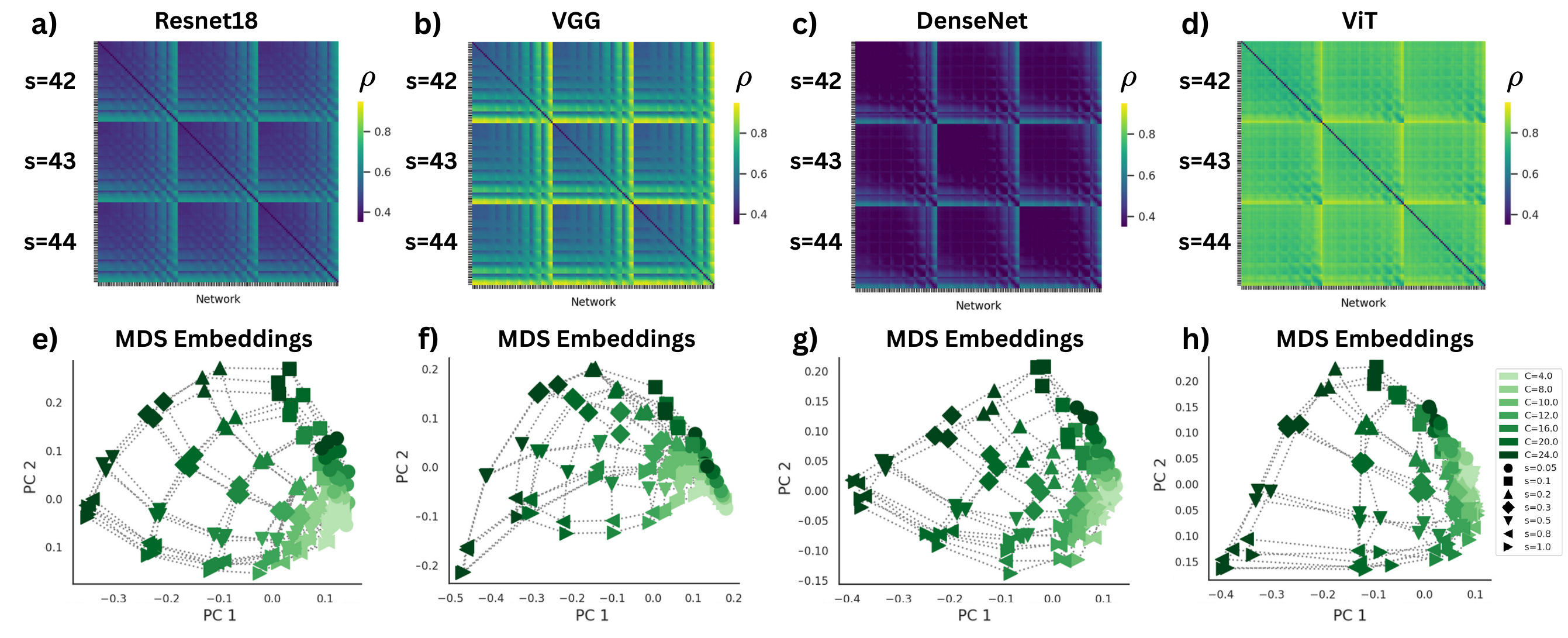} %
  \caption{Replicating the experiment of \Cref{fig:random_seeds} in four different architectures (Resnet18, VGG, DenseNet, ViT) with data augmentation method patch Gaussian. }\label{fig:seeds}
\end{figure}

\begin{figure}[h]
    \centering    
    \includegraphics[width=\linewidth]{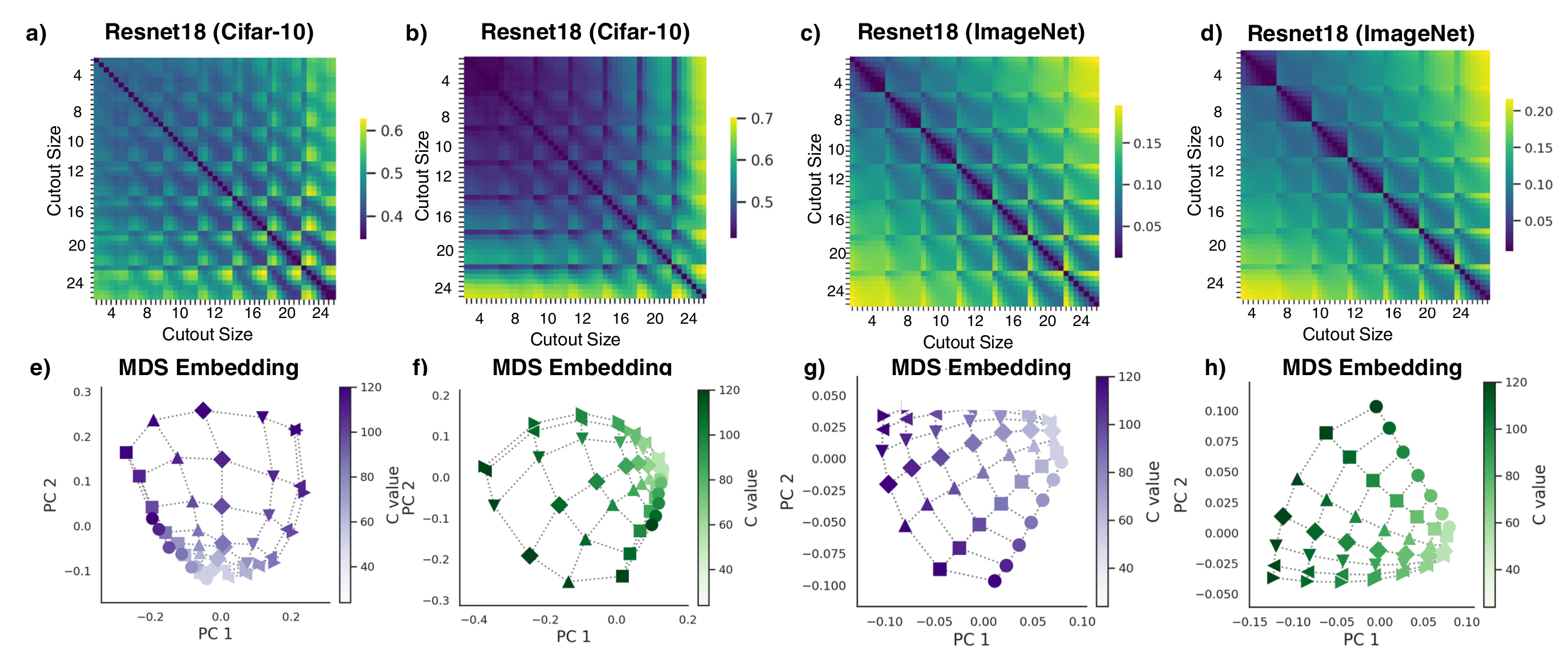} %
  \caption{Replication of \Cref{fig:fig2} using ResNet18 on CIFAR-10 and ImageNet. Panels (a,b) show models trained on CIFAR-10, while (c,d) show pretrained ResNet18 fine-tuned on ImageNet. The top row shows pairwise distance matrices across Cutout sizes, measured at Layer 2.1 for (a,c) and avgpool for (b,d). The bottom row (e–h) shows the corresponding MDS embeddings after PCA, colored by Cutout size. }\label{fig:wideresnet}
\end{figure}

\begin{figure}[h]
    \centering    
    \includegraphics[width=\linewidth]{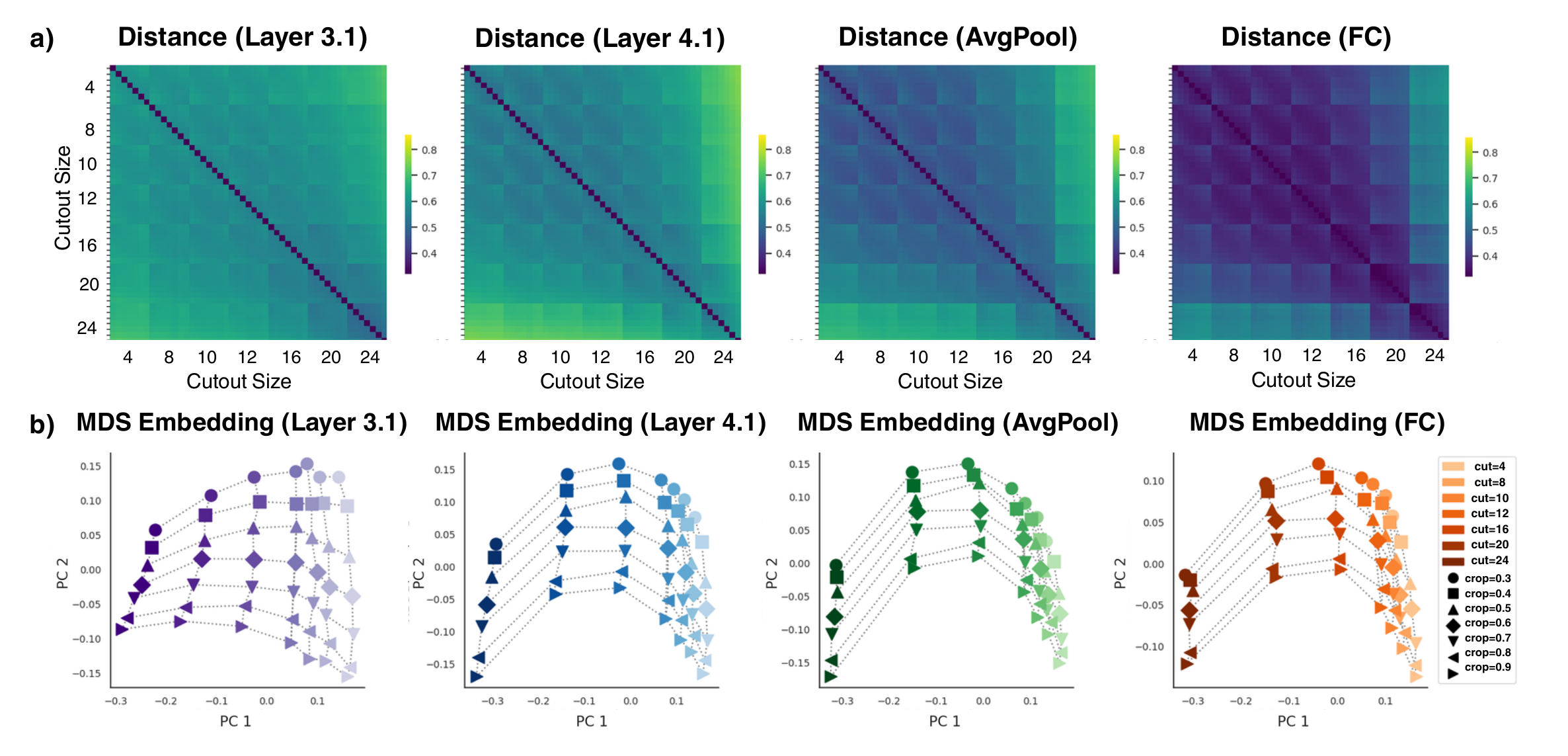} %
  \caption{Replicating the experiment of \Cref{fig:fig2} also in ResNet18 but using the data augmentation methods CutOut and random cropping. (a) is the distance metrics across pairse of augmentations, (b) is the MDS embeddings colored by augmentation strength. }\label{fig:cutout_crop}
\end{figure}

\begin{figure}[h]
    \centering    
    \includegraphics[width=\linewidth]{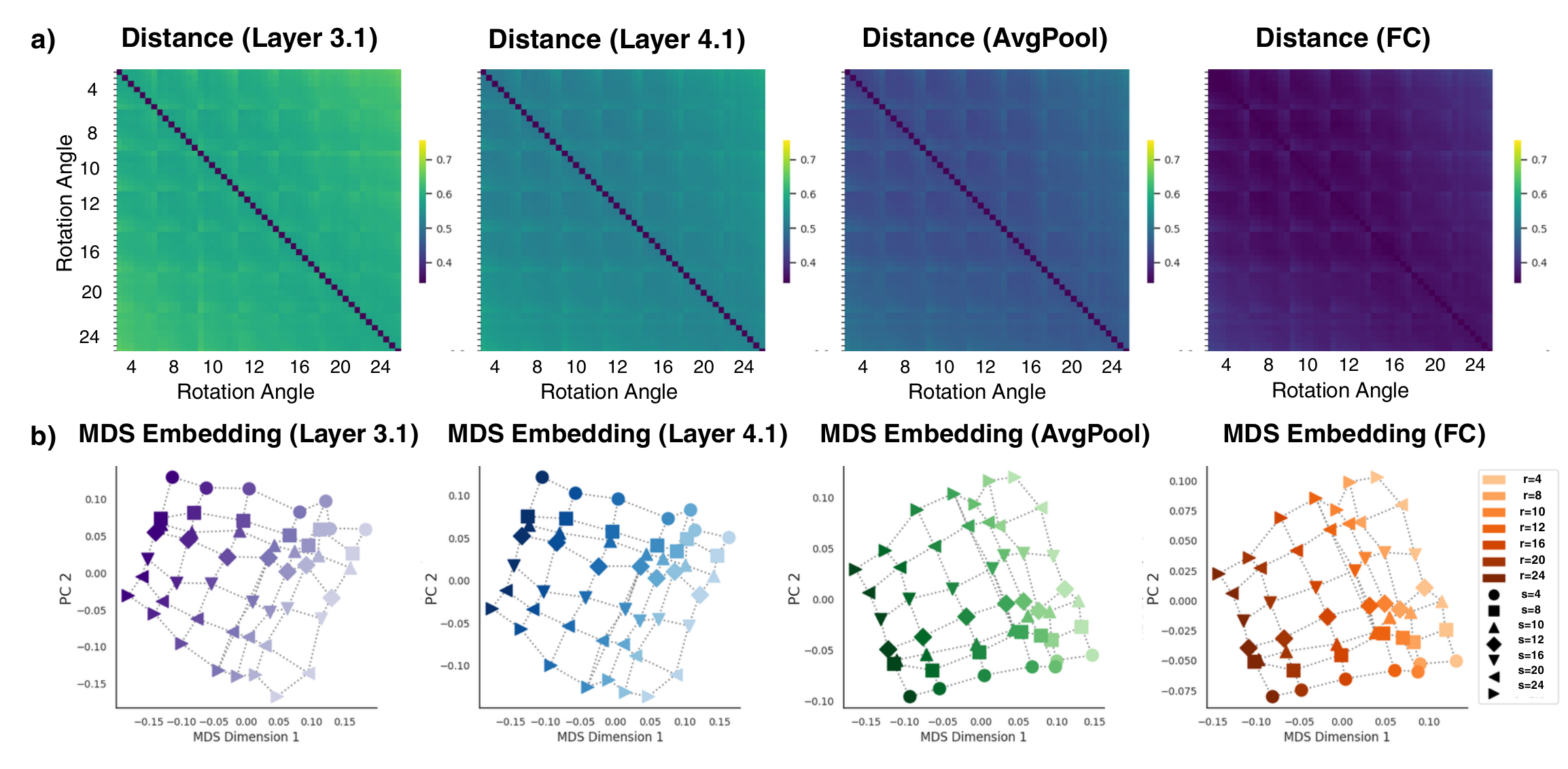} %
  \caption{Replicating the experiment of \Cref{fig:fig2} also in ResNet18 but using the data augmentation methods rotation and shear. (a) is the distance metrics across pairse of augmentations, (b) is the MDS embeddings colored by augmentation strength.}\label{fig:rot_sheer}
\end{figure}

To ensure the generalization and reproducibility of our results, we repeat the same experiment over different architectures (Fig \ref{fig:seeds}), different dataset (Fig \ref{fig:wideresnet}), and different augmentation types (Fig \ref{fig:cutout_crop}, Fig \ref{fig:rot_sheer}). We can observe similar structured geometry in shape space.

\begin{figure}[h]
    \centering    
    \includegraphics[width=1.2\linewidth]{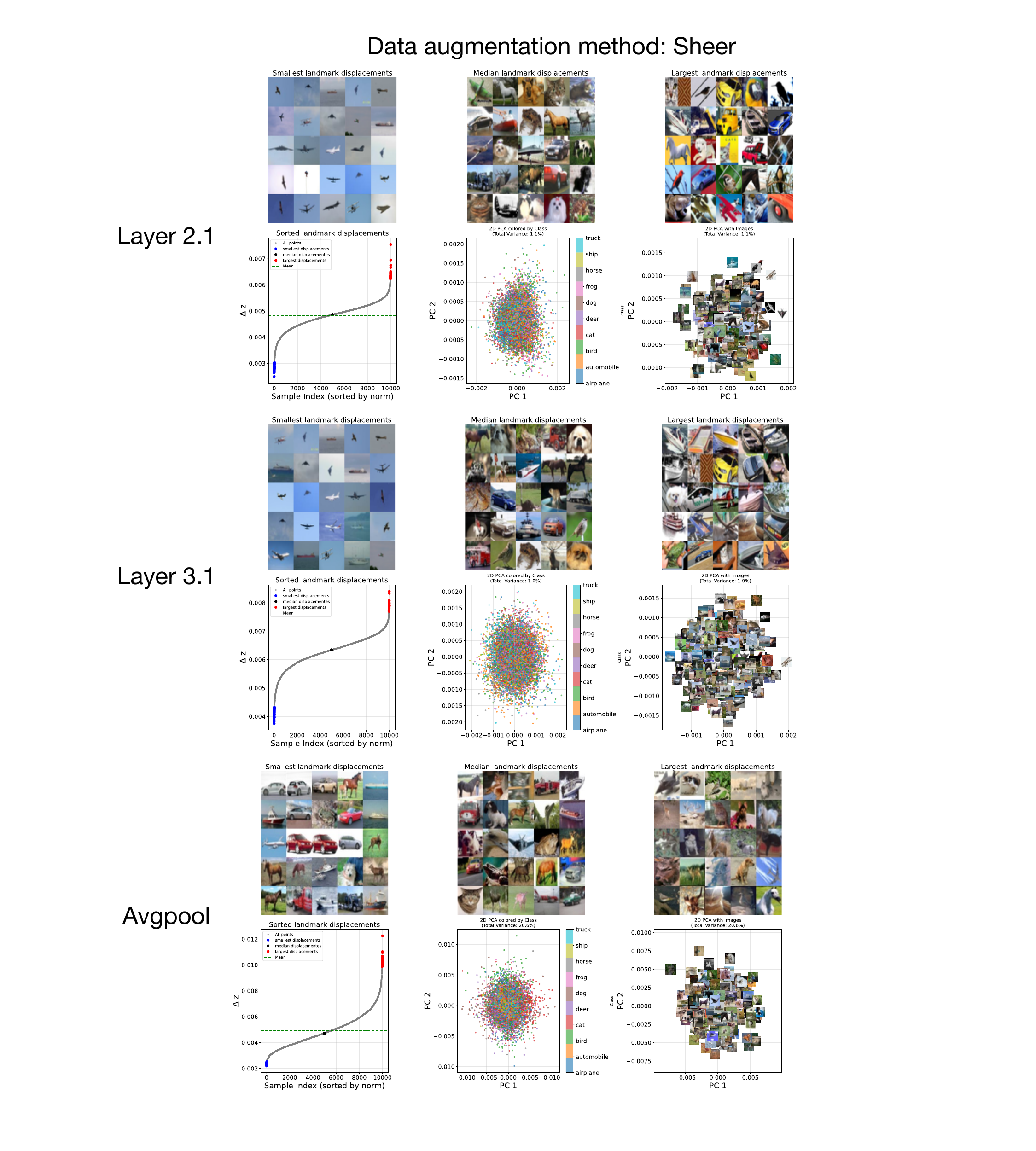} %
  \caption{Layer-by-layer variability in maximally, median,  and minimally deformed landmarks.  Landmark displacement analysis as in \Cref{fig:maxminimages} comparing layers of ResNet18 trained on un-augmented images and trained using data augmentation method sheer at level s = 4 (one step along the sheer trajectory in \Cref{fig:single_augments} a). PCA plots in bottom row are identical projections of the displacement $\Delta Z$ onto its first two principal axes, except one is colored by CIFAR-10 class and the other displays the corresponding CIFAR-10 image over each datapoint.}\label{fig:sheer}
\end{figure}

\begin{figure}[h]
    \centering    
    \includegraphics[width=1.2\linewidth]{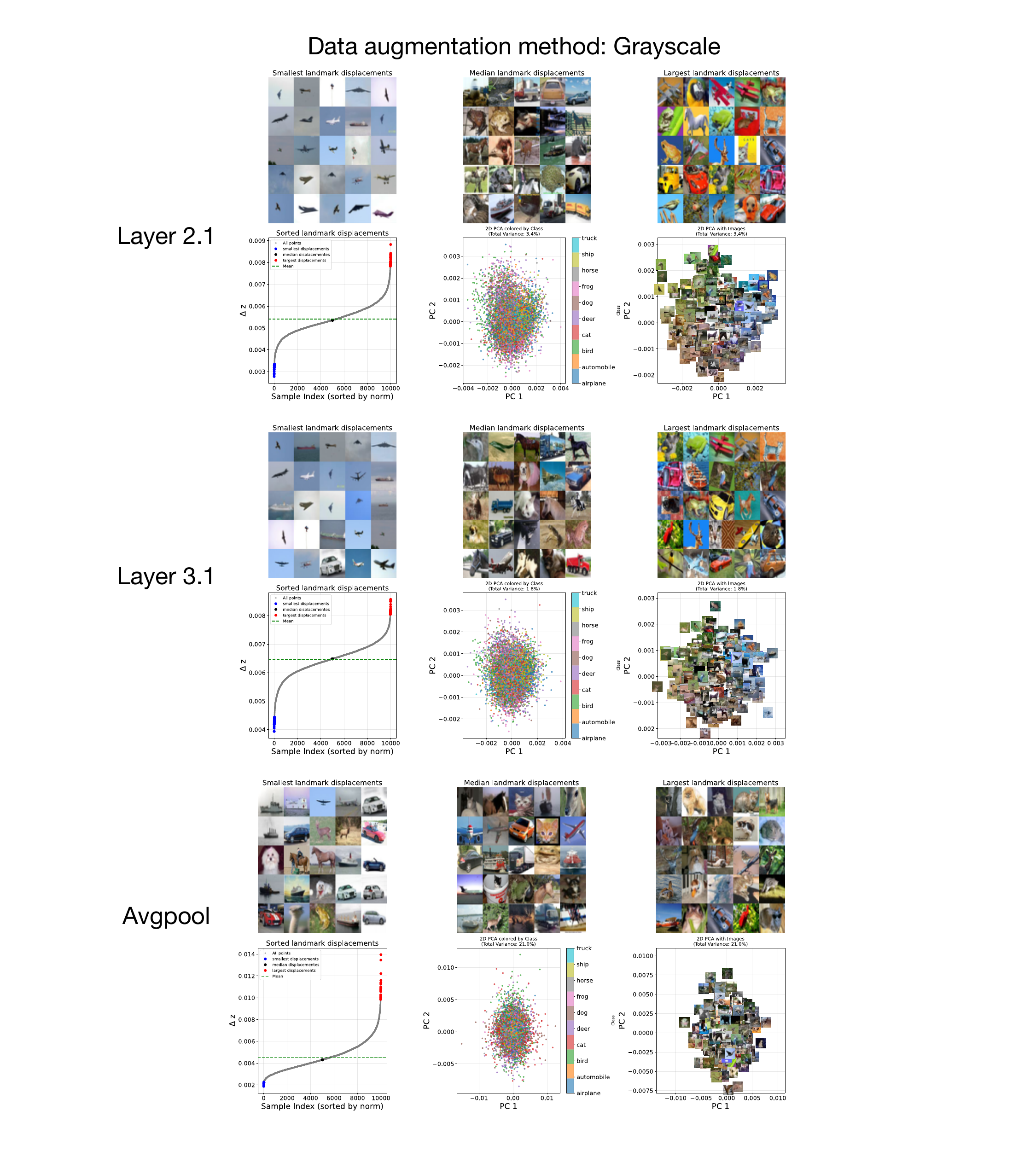} %
  \caption{ Layer-by-layer variability in maximally, median, and minimally deformed landmarks.  Landmark displacement analysis as in \Cref{fig:maxminimages} comparing layers of ResNet18 trained on un-augmented images and trained using data augmentation method `grayscale', with probability of image grayscale = 0.1 (one step along the grayscale trajectory of \Cref{fig:single_augments}). PCA plots in bottom row are identical projections of the displacement $\Delta Z$ onto its first two principal axes, except one is colored by CIFAR-10 class and the other displays the corresponding CIFAR-10 image over each datapoint.}\label{fig:grayscale}
\end{figure}

\begin{figure}[h]
    \centering    
    \includegraphics[width=1.2\linewidth]{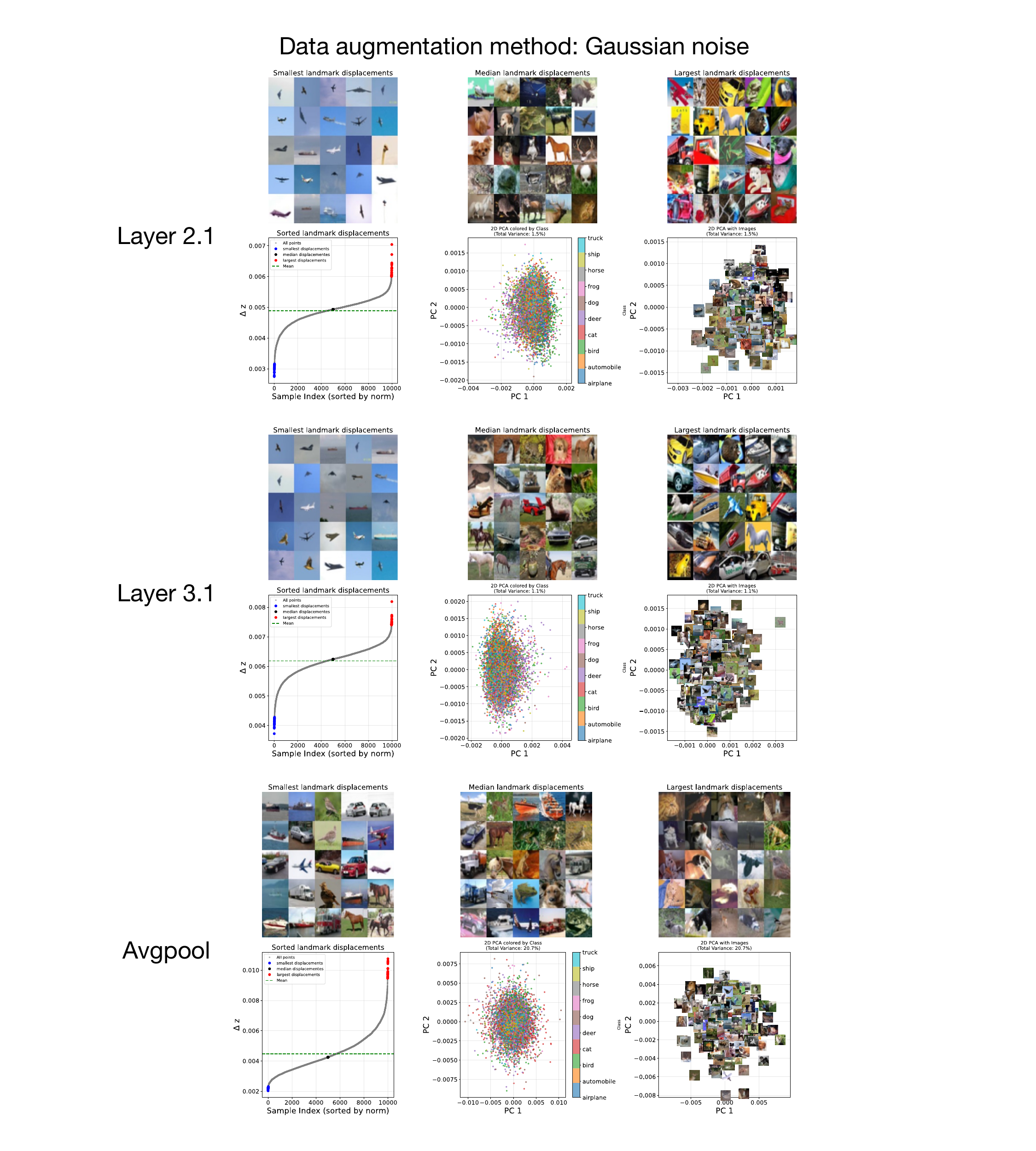} %
  \caption{ Layer-by-layer variability in maximally, median, and minimally deformed landmarks.  Landmark displacement analysis as in \Cref{fig:maxminimages} comparing layers of  ResNet18 trained on un-augmented images and trained using data augmentation method Gaussian noise at level s = 0.02.  PCA plots in bottom row are identical projections of the displacement $\Delta Z$ onto its first two principal axes, except one is colored by CIFAR-10 class and the other displays the corresponding CIFAR-10 image over each datapoint.}\label{fig:gaussian}
\end{figure}



\clearpage
\newpage

\end{document}